%% file: root.tex
\documentclass[letterpaper, 10 pt, conference]{ieeeconf}  
\IEEEoverridecommandlockouts                             
\usepackage{amsmath}
\usepackage{amsfonts} 
\usepackage{graphicx} 
\usepackage[labelfont=bf]{caption}
\usepackage{subcaption}
\usepackage{booktabs}
\usepackage{epstopdf}
\usepackage{multirow}
\usepackage{booktabs}
\usepackage{verbatim}
\usepackage[noadjust]{cite}
\usepackage{adjustbox}

\DeclareMathOperator*{\argmax}{arg\,max}
\DeclareMathOperator*{\argmin}{arg\,min}
\title{\LARGE \bf
Adversarial Attacks on Optimization based Planners
}

\author{Sai Vemprala and Ashish Kapoor$^{1}$
\thanks{$^{1}$Sai Vemprala and Ashish Kapoor are with Microsoft Corporation, Redmond, USA. \texttt{Sai.Vemprala@microsoft.com, akapoor@microsoft.com}}}%

\begin{document}

\maketitle
\thispagestyle{empty}
\pagestyle{empty}

\begin{abstract}
    Trajectory planning is a key piece in the algorithmic architecture of a robot. Trajectory planners typically use iterative optimization schemes for generating smooth trajectories that avoid collisions and are optimal for tracking given the robot's physical specifications. Starting from an initial estimate, the planners iteratively refine the solution so as to satisfy the desired constraints. In this paper, we show that such iterative optimization based planners can be vulnerable to adversarial attacks that force the planner either to fail completely, or significantly increase the time required to find a solution. The key insight here is that an adversary in the environment can directly affect the optimization cost function of a planner. We demonstrate how the adversary can adjust its own state configurations to result in poorly conditioned eigenstructure of the objective leading to failures. We apply our method against two state of the art trajectory planners and demonstrate that an adversary can consistently exploit certain weaknesses of an iterative optimization scheme.
\end{abstract}

\input{sections/intro}

\input{sections/related}

\input{sections/approach}
\input{sections/impl}
\input{sections/results}

\input{sections/conclusions}

\bibliographystyle{IEEEtran}
\bibliography{IEEEabrv, root}

\end{document}

%% file: sections/intro.tex
\section{Introduction}
	Trajectory planning is critical to build robots that perform desired tasks, where the goal is to determine a sequence of state configurations that is dynamically feasible and adheres to environmental constraints. Such constraints include avoiding undesired contact with themselves (self-collisions) and the environment (collision with obstacles or other robots). Given the central role of trajectory planners in robots proximal to humans or other equipment, it is important to consider their safety properties and failure modes. 
	In this paper, we explore and discuss adversarial attacks that can target trajectory planners. Specifically, we focus on the class of planners that determine valid solutions via minimization of a cost function \cite{ratliff2009chomp, Oleynikova2016, zhou2019robust}. The cost function mathematically encodes preference for dynamically feasible trajectories that help solve the task along with environmental and collision avoidance constraints. We build this work upon the key insight that an adversary has a control over the cost function being optimized by the planner, especially through the collision avoidance constraints. Thus, an adversary can force the planner to minimize cost functions that have ill-conditioned properties leading to failures. Examples of ill-conditioned properties include poor eigenvalue structure or small gradients that cause the optimizer to take a long time to converge. A poor eigenvalue structure can lead to slow convergence as the magnitudes of the minimum and maximum eigen values of a problem determine the speed of convergence of gradient descent. Long convergence times of planners are problematic as a valid actionable solution might not be available in real time. Fig 1. demonstrate how hardness of cost function changes with problem configurations. 
	
\begin{figure}[!tbp]
\centering
    \includegraphics[width=0.5\textwidth]{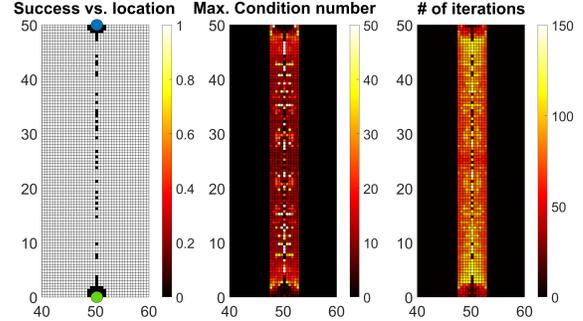}
    \caption{The success of a trajectory planner can be affected by problem configuration. We consider a simple 2-dimensional trajectory planning problem with gradient based optimization, with a robot attempting to plan a path from the green circle to the blue circle (left). Each cell in the grid maps to the success/failure of the planner if a single obstacle were present at that cell. Figure 1(left) shows that the planner can fail even in this simple scenario due to ambiguous gradients (black indicating obstacle positions causing failure). Figure 1(center) and (right) show that some obstacle locations, although the optimization is successful, cause complex configurations that are harder to solve - which is reflected either in a high condition number (center) (see section 3) or the optimizer requiring a large number of iterations to converge (right).}
    \label{fig:idea}
\end{figure}

	As the state of an adversary is part of the optimization problem/cost function being minimized continuously by the trajectory planner, we show that optimization based planners can be made susceptible to adversarial attacks. An adversary can reason and take actions such that the problem structure of the target's trajectory optimization is worsened. We discuss the mathematical background of optimization schemes that allows for this vulnerability to be exploited, and demonstrate that through this adversarial attack, it is possible to force iterative optimization based planners to endure an ill-posed or hard to solve problem, resulting in the optimizer taking more iterations to solve the problem, or in complex situations, fail altogether. Furthermore, we show that this is feasible as a black-box adversarial attack, where the adversary does not have access to any internals of the target's optimization scheme, and operates only upon feedback of the target's movements. We achieve this by approximating the behavior of a target optimization based planner through a surrogate function, which is modeled using a Gaussian process prior. As the general behavior of most collision avoidance algorithms is similar, this can be applied to several planners. Bayesian optimization is used to maximize this function and thereby conduct adversarial attacks, but for efficiency, we also present a simpler heuristic that approximates similar behavior. We test our approach on two state of the art planners and show that simple adversarial attacks result in failures for multiple optimization methods. In summary, the core contributions of this work are as follows:
	\begin{itemize}
	    \item Formulation of adversarial attacks that target optimization-based planners by affecting the objective function.
	    \item An algorithm for black-box adversarial attacks on collision avoidance planners based on Bayesian Optimization. 
	    \item Experiments that explore such adversarial attacks and their effects on planning success and performance, tested on two state of the art planners.
	\end{itemize}

%% file: sections/related.tex
\section{Related Work}
\label{sec:background}
    While robot safety is an area getting increasing attention, most of the work on adversarial attacks in machine learning and autonomous systems have been limited to perception systems. Specifically, several adversarial examples have been designed that break neural networks into classifying objects wrongly through imperceptible image perturbations \cite{goodfellow2014explaining, szegedy2013intriguing}, or through universal patches placed in the scenes being imaged \cite{brown2017adversarial, lee2019physical}. Various real life safety implications of these attacks were identified such as attacks on stop signs affecting autonomous vehicles \cite{eykholt2018robust}, attacks on face detection algorithms \cite{sharif2016accessorize} etc. 
    
    Some parallels can also be drawn between the proposed idea and the ideas of resource exhaustion attacks, or denial of service attacks in computer security \cite{601330}. Such attacks aim to attack software or web services in a way that increases execution/response times, and in worst case scenarios, result in crashes. Recently, machine learning methods have been leveraged to detect such attacks \cite{elsayed2020ddosnet, doshi2018machine, petsios2017slowfuzz}.
    
    In the realm of safe planning, there are methods that explore worst case safety guarantees for controllers via constructs such as Safety Barrier Certificates \cite{luo2019multi, ames2016control, wang2017safety} and reachability analysis \cite{herbert2017fastrack, bajcsy2019efficient}. Additional approaches to controller safety include formal methods based strategies inspired by temporal logic \cite{sadigh2015safe, dey2016fast} and chance constraints \cite{ulusoy2014incremental}. Trajectory generation was also handled by enhanced sampling-based planners with safety certificates \cite{axelrod2018provably} and `funnel libraries' based on Lyupanov functions \cite{majumdar2017funnel}.
    
    Our adversarial attacks focus on planners using gradient-based optimization. CHOMP \cite{ratliff2009chomp} was one of the first planners that showed that trajectory optimization can be feasibly applied to path planning, through precomputed distance functions and gradient based optimization. STOMP \cite{kalakrishnan2011stomp} is another planner that first samples various candidate trajectories and minimizes a cost to converge onto a best-performing linear combination of the trajectories. Loco planner \cite{Oleynikova2016} presents a continuous-time trajectory optimization method inspired by CHOMP, adapting directly to quadrotor platforms by generating smooth polynomial continuous time representations. Loco planner is also part of a full mapping and replanning framework where trajectory optimization is performed on an initial estimate from an Informed RRT* algorithm \cite{oleynikova2018safe}. Fast planner \cite{zhou2019robust} is another continuous-time variant, which uses a coarse path plan returned by a kinodynamic A* and optimizes it into a smooth, collision-freeavoiding trajectory through B-spline optimization. FASTER \cite{tordesillas2020faster} uses a mixed integer QP formulation to solve trajectory planning in unknown environments.

%% file: sections/approach.tex
\section{Approach}
\label{sec:approach}

\subsection{Background: Optimization based planning}

Given a start and a goal state, a trajectory planner connects them through a set of configurations that are dynamically feasible, respect environmental constraints and are safely traversable by the robot. To achieve this through optimization, an objective function is created that codifies a specification of a goal, and safety requirements such as collision thresholds. Environmental factors and dynamic constraints such as velocity/acceleration limits are either expressed directly, making it a constrained optimization problem, or can be validated in a second phase after a collision-free trajectory is computed. Some trajectory optimization approaches \cite{Oleynikova2016, richter2016polynomial} formulate the trajectory through its endpoint derivatives rather than polynomial coefficients, thus solving trajectory optimization as an unconstrained quadratic program, which is faster to solve than the constrained variant. Formally, such trajectory planners solve an optimization problem of the following form:
\begin{equation}
    x^* = \argmin_x w_d J_d(x) + w_c J_c(x, a).
\end{equation}
Here $x$ refers to the sequence of parameters that determine the robot trajectory, such as the velocities and the positions, and are optimized through the planner. On the other hand $a$ corresponds to environment specific terms such as positions of obstacles, over which the robot has no influence.

The cost function is often decomposed into two main terms: the first part $J_d(x)$ models the desired characteristics of the solution. For example, it might be desirable to to minimize jerk to ensure the trajectory is smooth. The second term $J_c(x, a)$ enforces collision avoidance and can also include environmental constraints such as power conservation. The value of the weights determine relative importance amongst the various terms. Examples of planners that follow this design include CHOMP and Loco \cite{ratliff2009chomp, Oleynikova2016}.  

Various factors such as the environmental constraints, system dynamics and the parameter space determine the hardness of the optimization problem. For instance, presence of multiple collision objects can lead to non-convexity. Similarly, for vehicles such as quadrotors, it is common to plan trajectories as higher order polynomials in order to ensure smoothness - leading to a higher dimensional optimization problem. The trajectory planning problem thus requires optimizing the above mentioned objective and can be solved by a nonlinear optimization solver. In general, methods such as Newton's method, which iteratively minimizes a quadratic approximation of the objective function $f$ around the current point of evaluation $x_k$, are used in planners. Most of these methods require the computation of a full Hessian matrix or rely on fast approximation of the inverse Hessian (e.g. BFGS) \cite{nocedal2006numerical}.

The Hessian matrix encodes the second derivatives of the objective function and determines the convergence properties of the optimizer. Specifically, a Hessian and its eigenvalues represent the local curvature of the objective function. This {\em eigenstructure} of the optimization problem is critical in determining the convergence characteristics and the minimization performance. During the descent towards the minimum, the step size is bounded by the largest eigenvalue of the Hessian, but the rate of convergence is determined by the smallest. Hence, a particular metric of interest is the condition number of the Hessian, which is defined as the ratio of the largest eigenvalue to the smallest. An optimization problem with a large condition number is known to be `ill-conditioned' and causes poor convergence properties, because the function contours are stretched out and the gradient-based descent steps tend to point towards non-optimal directions. Other elements in the problem configuration can also have adverse effects, such as when the gradients are ambiguous around certain values (see Figure \ref{fig:idea}), or too small, resulting in an ineffective step size.

\subsection{Adversarial attacks}
Given that the optimization based planners rely on gradient-based iterative solvers, this allows one to create attacks such that the objective function becomes difficult to minimize. The key idea of this paper is to determine adversarial configurations, which directly influence the cost function and degrade the convergence characteristics of the optimizer.

Let us consider a scenario where a `target' robot is planning a trajectory (parameterized as $x$) towards a destination while avoiding obstacles (parameterized as $a$) through the cost function $f(x, a)\stackrel{\mathrm{def}}{=} w_d J_d(x) + w_c J_c(x, a)$. We assume that the map is known apriori, i.e., the states of all the obstacles are fully observable. An adversarial robot present in the same environment acts as another obstacle for the target robot; and any change in its state affects $a$, thus changing the cost function $f(x,a)$. The adversary can leverage this effect to change $a$ strategically to make the optimization problem ill-conditioned, hard to solve, or introduce a bad local minimum. While $a$ directly only forms a part of $J_c$, the resulting trajectory while minimizing this adversarially enhanced collision cost has a direct effect on $J_d(x)$. This can potentially create an imbalance in the problem structure. Bad local minima can also arise from adversarial states that are environmentally-aware, for instance, it is possible to trap the robot between multiple obstacles through an attack configuration, which generates ambiguous gradients. 

An adversarial attack constitutes determining a configuration $a^*$ that leads to poor convergence of the optimizer. For most of the popular optimization methods, including gradient descent, Newton’s method, BFGS, stochastic gradient descent, it generally holds that a higher condition number of the Hessian in the vicinity of the local optimum leads to poor convergence characteristics \cite{bertsekas1997nonlinear, nesterov2018lectures, bottou2018optimization}. Thus, the goal of the adversary is to find such environment configurations $a$ leading to a high condition number. Formally, we write this as:
\begin{equation}
    a^* = \argmax_a \kappa(a),
\end{equation}
where $\kappa(a)$ is the ratio of the largest eigenvalue of the Hessian to its smallest, computed at the optimum for the configuration $a$. Intuitively, a large condition number corresponds to a function whose value is sensitive to the perturbations in the input. 

Note that it is rare that the internals of the target's optimization scheme such as the cost function $f(x,a)$ would be accessible to an adversary, making it impossible for an adversary to compute $\kappa(a)$. Hence, it is necessary to design a black-box, generalizable method for adversarial attacks.

\subsection{Black Box Attacks}
We propose black box attacks via a surrogate function that models 
$\kappa(a)$ using observations of the target. Intuitively, any change in the trajectory of the target robot $r$ is primarily due to the effect of environment configuration parameter $a$. In a collision avoidance scenario, how a robot navigates around an obstacle provides information about the trajectory planner. Consequently, we use the angle between the observed velocity $\vec{v}_r(a)$ of the target due to the configuration $a$ from the nominal velocity $\vec{v}_r^0$ as a surrogate:
\begin{equation}
    \angle(\vec{v}_r(a), \vec{v}^0_r) \approx \kappa(a)
\end{equation}
The key intuition here is that this deviation in the robot's velocity is indicative of the cost landscape for objective $J_c(x,a)$ in eq. 1. Finding configurations $a$ that maximize this surrogate function corresponds to finding situations where the trajectory planner has to significantly deviate from a nominal solution. Large deviations in solutions are indicative that small perturbations in $a$ will lead to large changes in the planner output. Such sensitivity to the input results indicates an ill conditioned optimization problem, thus the goal of the adversary is to find configurations with maximal deviations.

We propose to carry out such an attack via Bayesian optimization \cite{mockus1978application}. Specifically, we first use Gaussian Processes (GP) \cite{rasmussen2003gaussian} to model the relationship between $\angle(\vec{v}_r(a), \vec{v}^0_r)$ and $a$. An initial GP model is created via a sequence of observations corresponding to the angle between two successive velocity vectors after one step of planning.

\begin{figure*}
\centering
    \begin{subfigure}[t]{0.34\textwidth}
        \hspace{-0.4cm}
      \includegraphics[width=\textwidth]{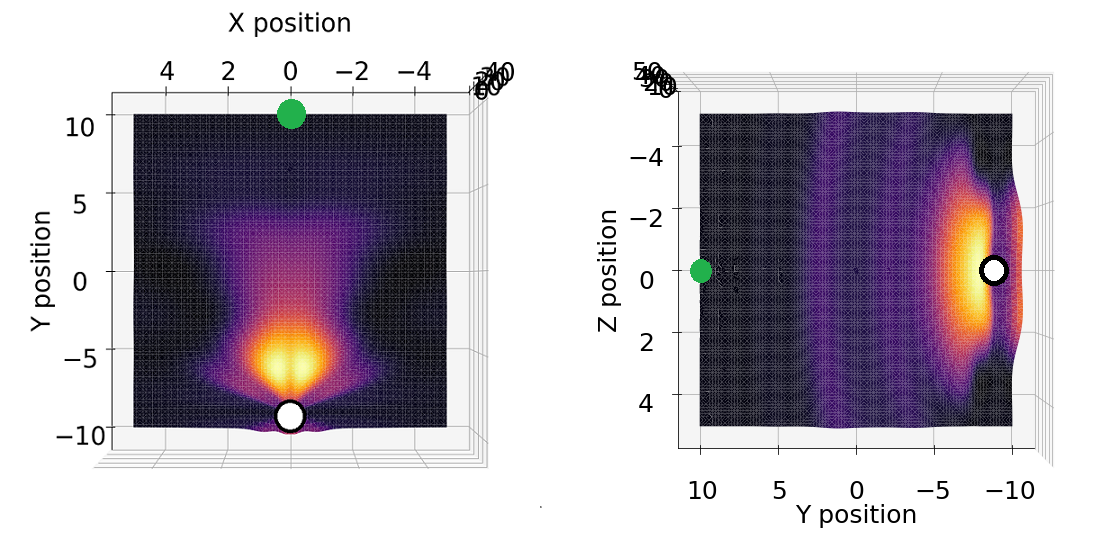}
      \caption{GP fit showing expected deviation in trajectory for adversary positions in X-Y. White circle indicates start position for target and green circle indicates goal.}
    \label{fig:gp}
    \end{subfigure}
    \begin{subfigure}[t]{0.28\textwidth}
      \includegraphics[width=\textwidth]{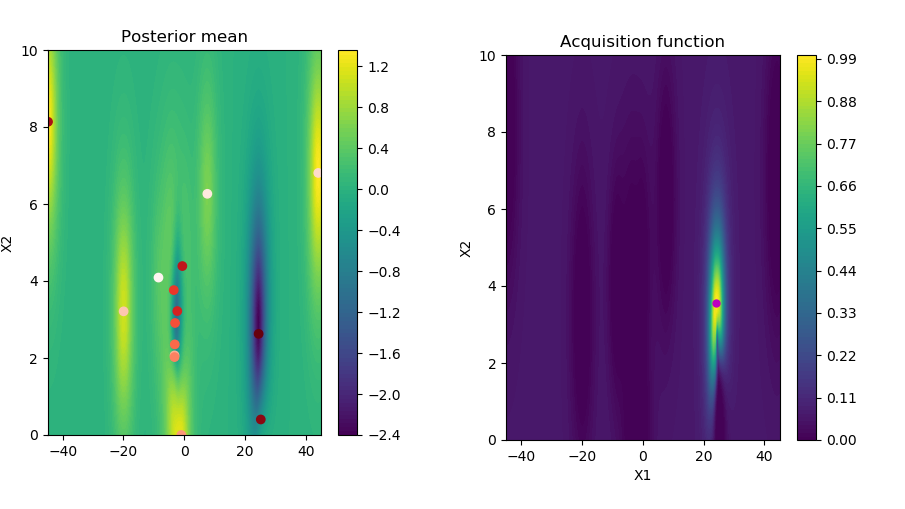}
      \caption{Example Bayesian optimization result for computing adversarial attack position. X1 and X2 represent $\theta$ and $r$ in spherical coordinates relative to target.}
    \label{fig:bo}
    \end{subfigure}
    \begin{subfigure}[t]{0.17\textwidth}
      \includegraphics[width=\textwidth]{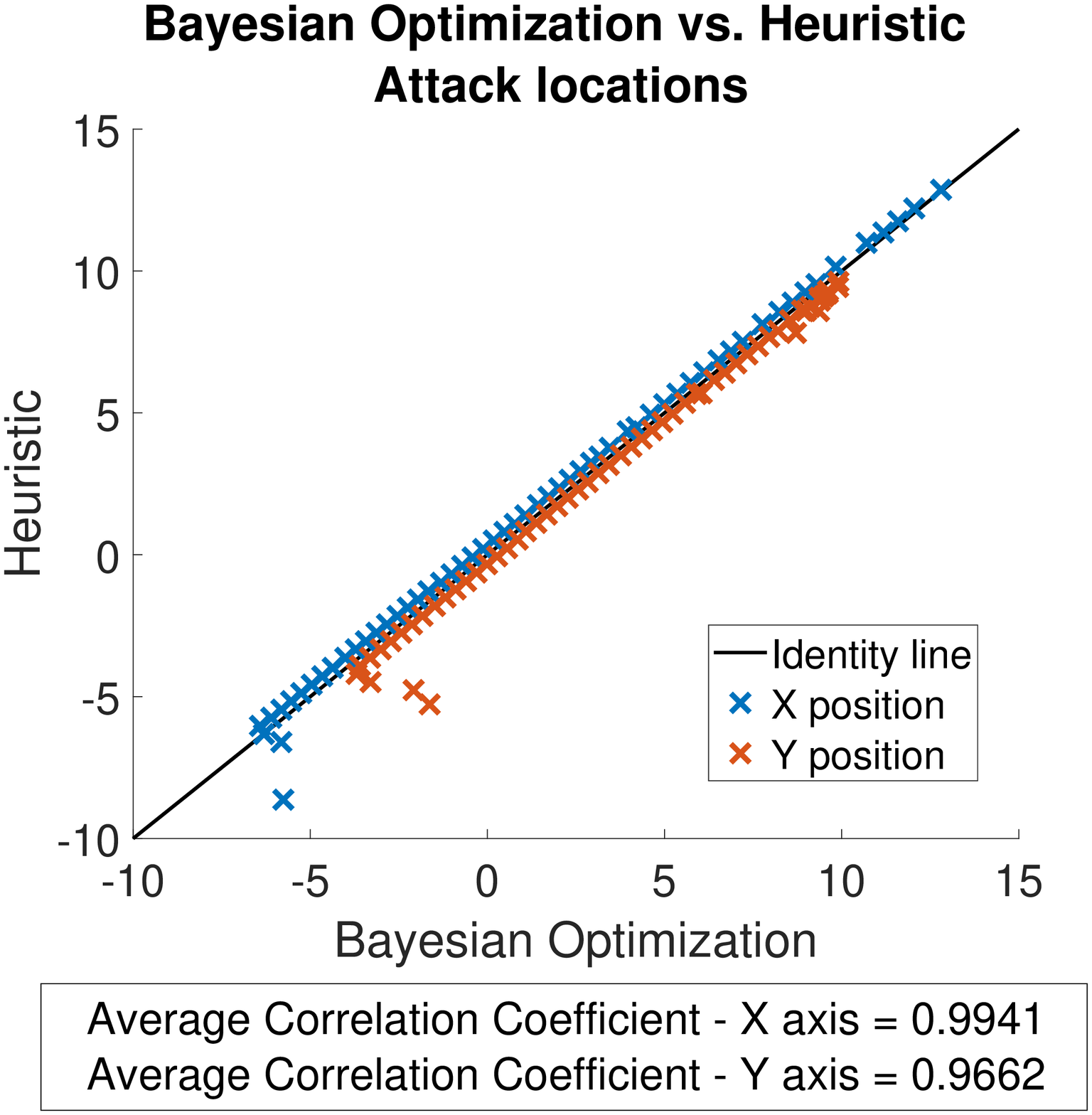}
      \caption{Strong correlation between outputs (XY positions) from Bayesian optimization and the heuristic}
    \label{fig:compare_bo_states}
    \end{subfigure}
    \begin{subfigure}[t]{0.17\textwidth}
      \includegraphics[width=\textwidth]{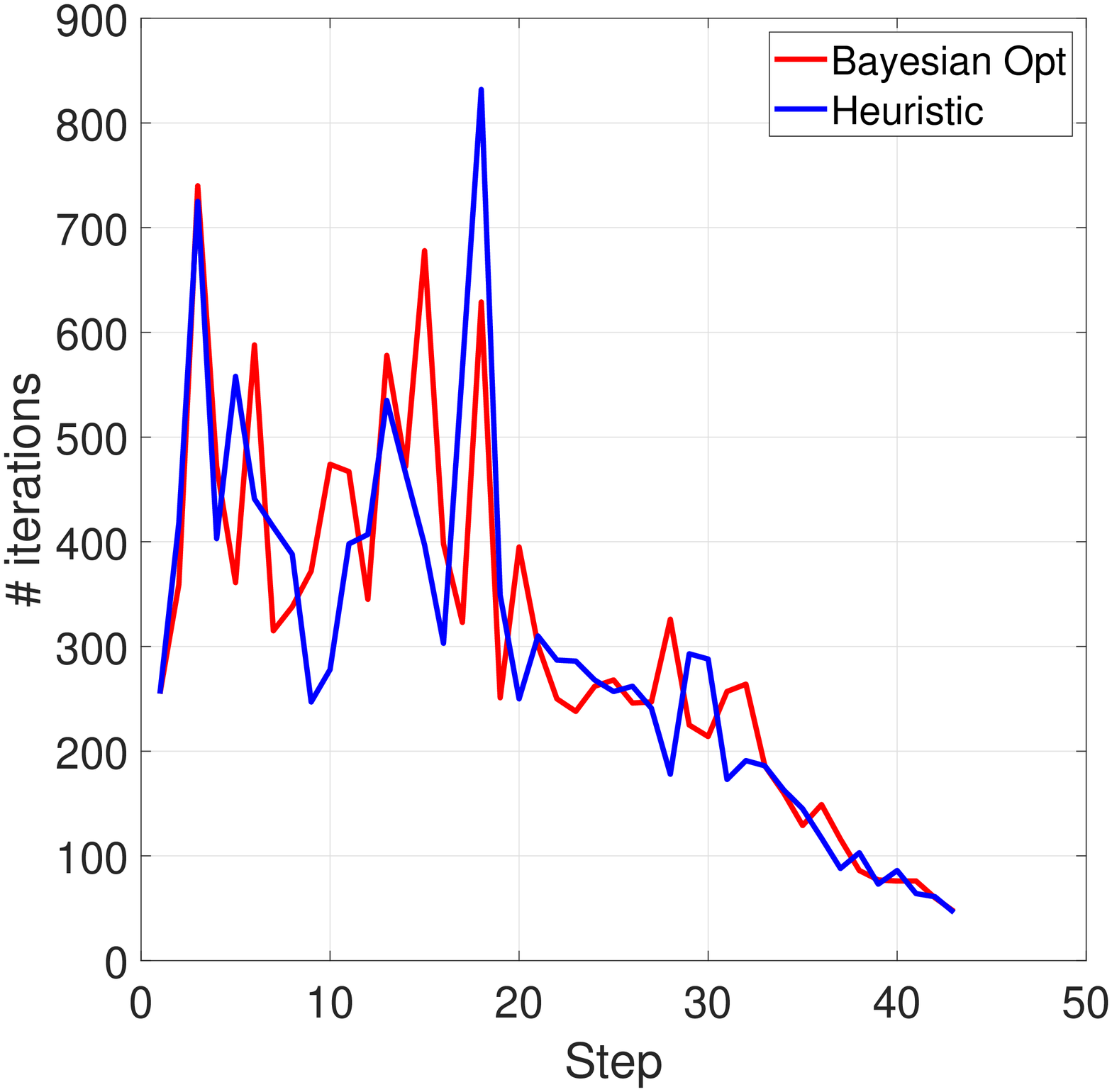}
      \caption{Comparable behavior in the target optimizer between BO and heuristic, over a single trial.}
    \label{fig:compare_bo_iters}
    \end{subfigure}
    \caption{Samples of Gaussian process modeling planner behavior and operation of the Bayesian optimization scheme over the GP prior, along with a comparison of the BO attack results and the ones from the heuristic approximation.}
\end{figure*}

Next, our goal is to find instances of $a$ such that the trajectory planner will output a solution with large deviation. Bayesian optimization accomplishes this task by exploiting the uncertainty around the GP predictions, and then efficiently exploring the parameter space to find promising candidates. In this work, we use a Radial Basis Function (RBF) kernel for GPs, and use Expected Improvement (EI) \cite{mockus1978application} as an acquisition function for Bayesian Optimization.

As Bayesian optimization is typically quite expensive to carry out in real time, we also propose an empirically found approximation that is computationally efficient. Specifically, the approximation entails computing a trajectory by the adversary to the following attack configuration:
\begin{equation}
\theta_a = \theta_r +  0.5\Delta t \frac{\partial \theta_r}{\partial t}
\end{equation}
Here $\theta_r$ is the most current observed configuration for the robot under attack, expressed as an angle of the velocity vector. $\Delta t$ is the time period between the planning cycles. The deviation term $\frac{\partial \theta_r}{\partial t}$ can be computed empirically via historical observations for the target robot, for example, by measurements of the position of the target. This approximation follows from the assumption, similar to the surrogate function, that any deviation in the target's trajectory over time is a result of an attempt for collision avoidance. Such deviations are thus informative about the inherent optimization objective being optimized. The approximation in eq. 4 aims to position the adversary on the opposite side of the cause of this deviation and attempts to create a problematic configuration in the cost landscape. Assuming $\frac{\partial \theta_r}{\partial t}$ to have a range of [$\frac{-\pi}{2}$, $\frac{\pi}{2}$], we empirically choose a scaling factor of $0.5$ so that the output of the heuristic matches the range of the Bayesian optimization outputs.

\begin{figure}
   
\end{figure}

As a black-box attack, our method makes no assumptions about the structure of the underlying Hessian matrix. The aim of this attack is to take a simpler planning problem that might contain a well-conditioned Hessian, and attempt to degrade the problem structure and make the matrix ill-conditioned.




%% file: sections/impl.tex
\section{Experiments}

\subsection{Implementation}

To model planner behavior as described above, we fit a GP using data from multiple trials with a random obstacle. An example fit can be seen in \ref{fig:gp}, the hotspots indicating high deviations in target velocity vector. Assuming full knowledge of the map and the target's state, this modeling can also be performed online. Bayesian Optimization was run over this GP for initial tests, an example computation of optimal attack location can be seen in \ref{fig:bo}. For the experiments, the heuristic in eq. 4 was used as a replacement for Bayesian optimization to compute adversarial states. Through empirical testing, we observe the states output by this heuristic to be close to the outputs of the Bayesian optimization scheme, and the effects on the optimizer to be similar (examples in \ref{fig:compare_bo_states}, \ref{fig:compare_bo_iters}). As mentioned in section 3, the heuristic was only used to replace the Bayesian Optimization for computational efficiency, the two are interchangeable.

We evaluate the effects of these adversarial attacks on two trajectory planning algorithms with collision avoidance: Loco planner \cite{Oleynikova2016}, and the Fast planner \cite{zhou2019robust}; both of which are aimed at quadrotor trajectory planning. While we do not use exact quadrotor dynamics models for the target or adversary, the use of these planners is to demonstrate the effect of other parameters (such as trajectory smoothness constraints) in the optimization scheme. We describe the problem setting for the implementation and the corresponding assumptions below.

\begin{enumerate}
    \item The target and the adversary share a common map with a non-zero number of environmental obstacles; and have full knowledge of each other's states and the map.
    \item The trajectories planned and tracked are only in the position space, and the orientations are ignored.
    \item The target is assumed to be moving at a constant velocity, and the adversary has a maximum velocity limit that that does not exceed that of the target. 
\end{enumerate}

An adversarial trial is defined as a sequence of steps, where the adversary continuously tries to position itself based on target state feedback. Replanning occurs at each step, and the resultant trajectory is sampled to obtain a new position for the target, and subsequently the adversary. An adversarial trial results in one of the following outcomes for the target: reaching the goal, collision with a map obstacle (Coll-M) or collision with the adversary (Coll-A). We also monitor the iteration count of the optimizer. A high number of iterations required to solve the problem, although planning is successful, is indicative of an adverse effect. In real time deployment, systems often have a hard requirement of having to compute a safe plan within a certain timeframe, which translates to a maximum iteration count.

\begin{figure}[!t]
\centering
  \begin{subfigure}[t]{0.22\textwidth}
    \includegraphics[width=\textwidth]{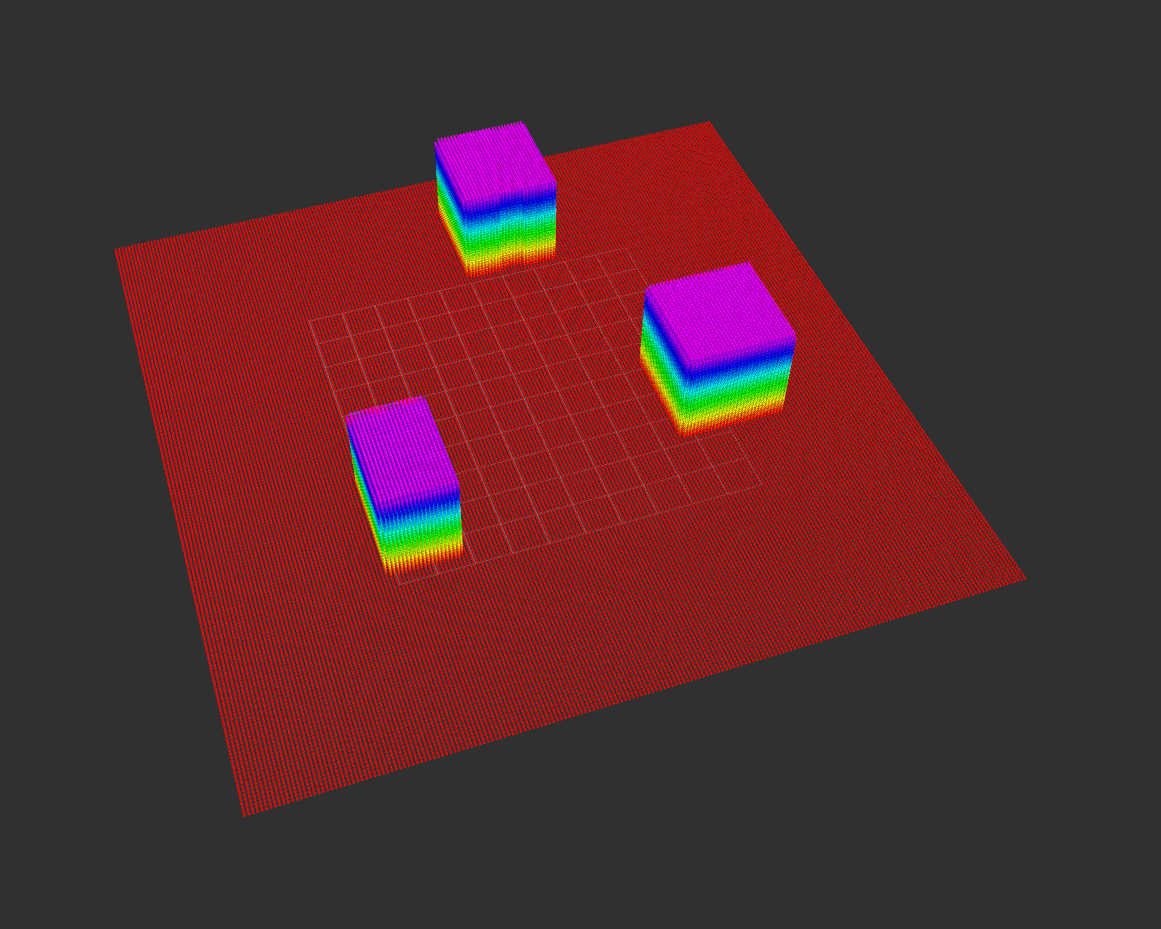}
    \caption{Sparse environment}
    \label{fig:sparsemap}
  \end{subfigure}
  \quad
  \begin{subfigure}[t]{0.11\textwidth}
    \includegraphics[width=\textwidth]{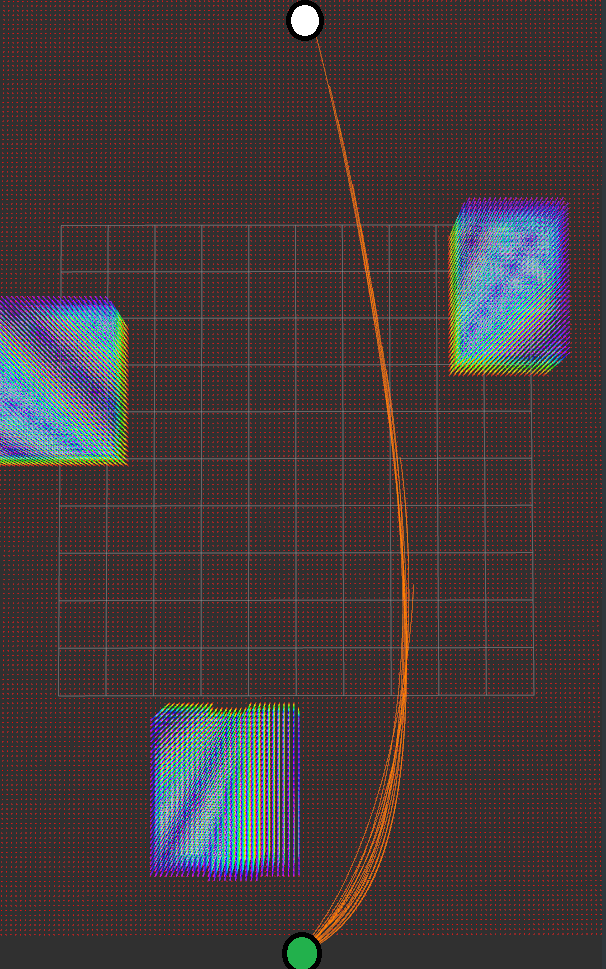}
    \caption{No adv.}
    \label{fig:normalplan}
  \end{subfigure}
  \quad
    \begin{subfigure}[t]{0.12\textwidth}
    \includegraphics[width=\textwidth]{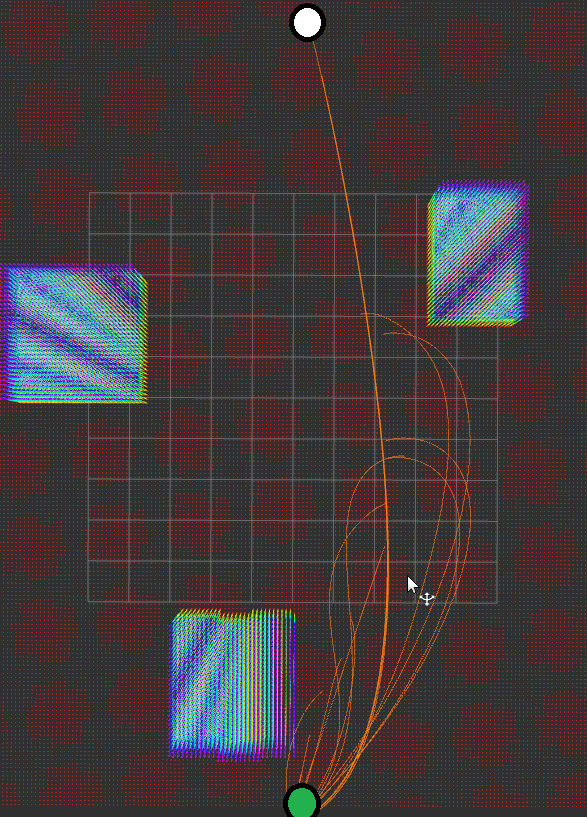}
    \caption{With adv.}
    \label{fig:advplan}
  \end{subfigure}
  \quad
  \begin{subfigure}[t]{0.22\textwidth}
    \includegraphics[width=\textwidth]{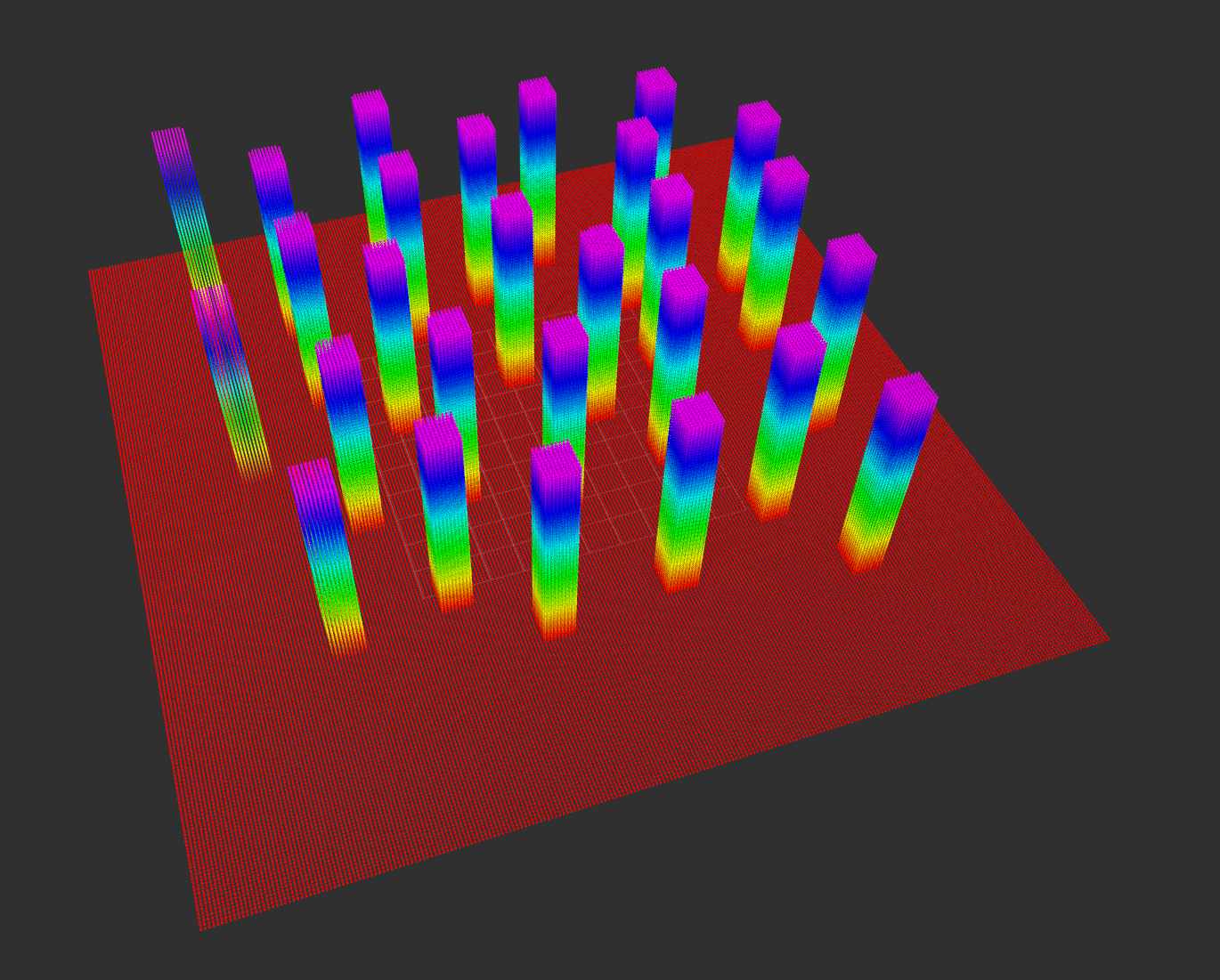}
    \caption{Dense environment}
    \label{fig:densemap}
  \end{subfigure}
  \caption{Environments used for testing: (a) and (d); and a visual depiction of the effect of an adversarial attack. (b) shows trajectories planned from start (white) to goal (green) normally, but if an adversary were continually present in adversarial states, the trajectories get into unsafe regions near the obstacles. (c)}
\end{figure}
The open sourced implementations of the Loco and Fast planners were utilized, which contain a BFGS solver for loco planner, L-BFGS and SLSQP versions for Fast planner. We test the adversarial attack scheme in two environments which refer to as \emph{sparse} (\ref{fig:sparsemap}), with a small number of obstacles; and \emph{dense} with a larger obstacle field (\ref{fig:densemap}). In both cases, the target robot spawns randomly at one edge of the map and is commanded to plan a path to the opposite side, while the adversary is spawned at a random location. For both planners, we test two configurations: \textit{default}, where the weights in the cost function were kept at their default values; and another as \textit{conservative}, where the weight on the collision cost was increased. We provide an Euclidean signed distance field \cite{oleynikova2016signed} map representation to the planners, and we use a simple repulsion method to avoid the adversary colliding with the obstacles. The adversary and the target are both in motion, with the rest of the map assumed to be static. 

\begin{figure*}[!t]
        \centering
        \begin{subfigure}[b]{0.24\textwidth}
            \centering
            \includegraphics[width=\textwidth]{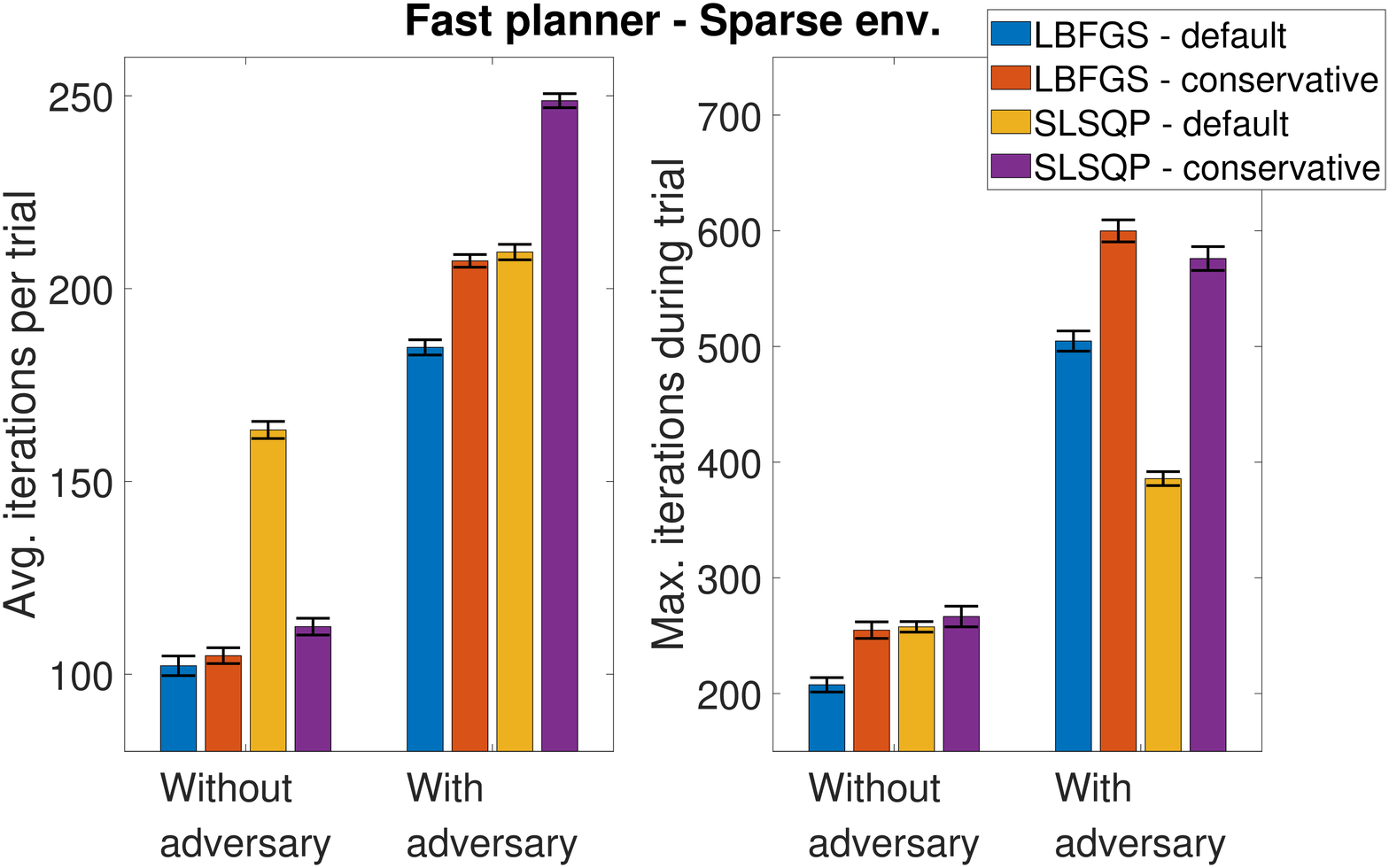}
            \caption{Fast planner - \textit{Sparse}}
            \label{fig:fastsparse}
        \end{subfigure}
        \begin{subfigure}[b]{0.24\textwidth}  
            \centering 
            \includegraphics[width=\textwidth]{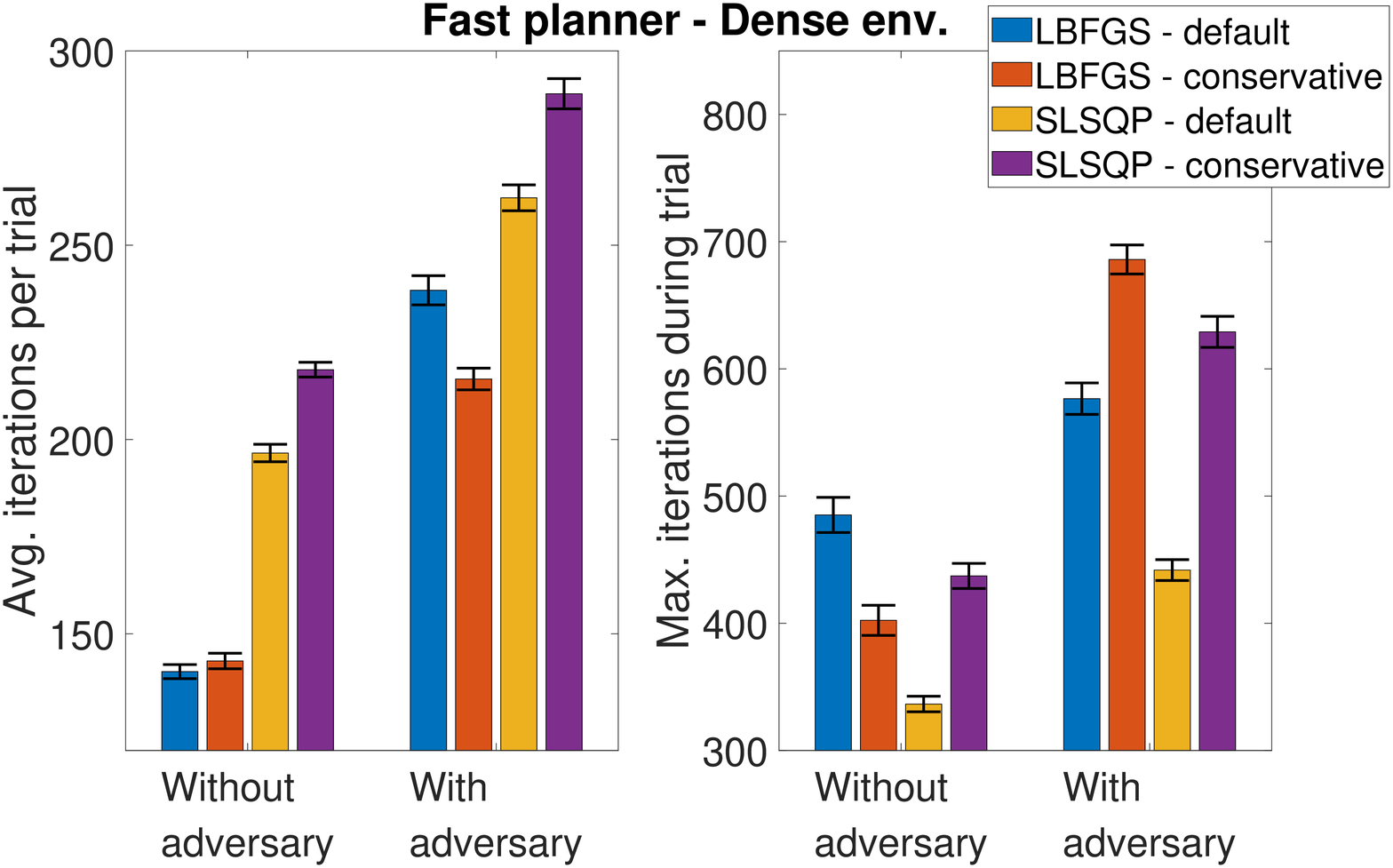}
            \caption{\small Fast planner - \textit{Dense}} 
            \label{fig:fastdense}
        \end{subfigure}
        \begin{subfigure}[b]{0.24\textwidth}   
            \centering 
            \includegraphics[width=\textwidth]{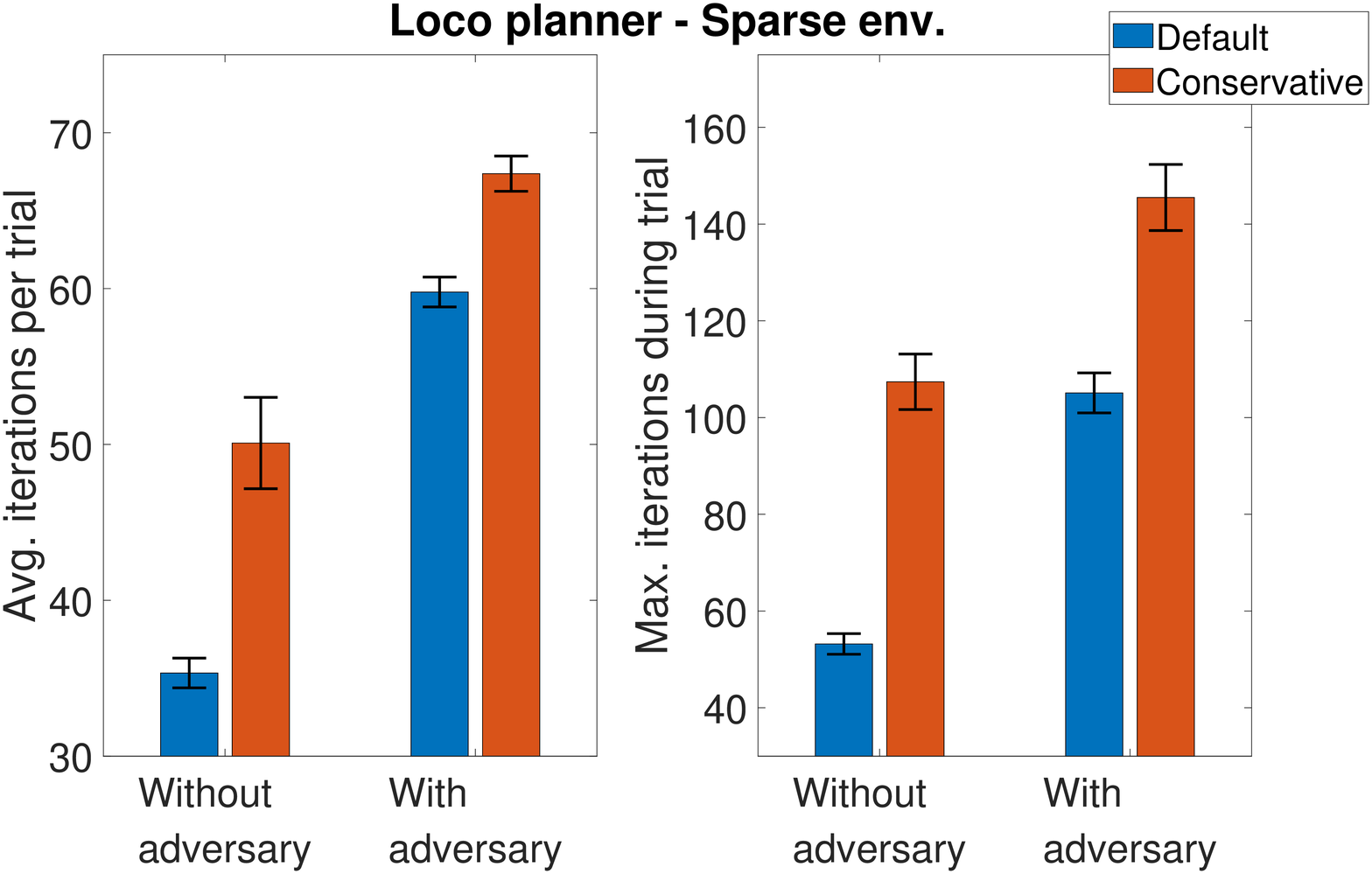}
            \caption{Loco planner - \textit{Sparse}}  
            \label{fig:locosparse}
        \end{subfigure}
        \begin{subfigure}[b]{0.24\textwidth}   
            \centering 
            \includegraphics[width=\textwidth]{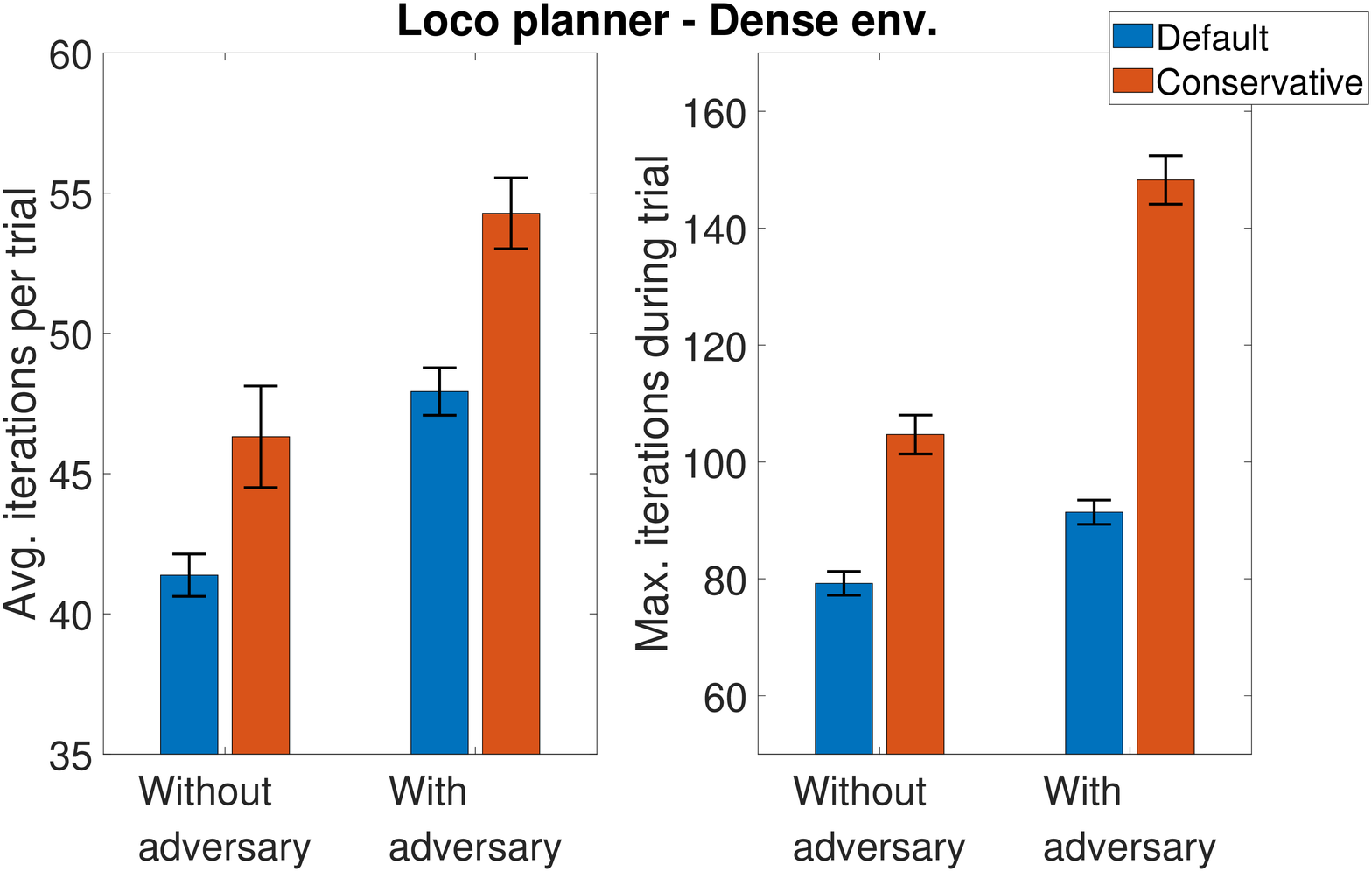}
            \caption{Loco planner - \textit{Dense}}
            \label{fig:locodense}
        \end{subfigure}
        \caption{Effect of adversarial attacks on average iteration count and maximum iteration count seen during trajectory optimization. Averaged over 100 trials for both Loco and Fast planners; standard error shown as error bars shows statistical significance.} 
        \label{fig:planners_itcount}
\end{figure*}

%% file: sections/results.tex
\subsection{Results}

\begin{figure*}[!tbp]
\centering
  \begin{subfigure}[t]{0.38\textwidth}
    \includegraphics[width=\textwidth]{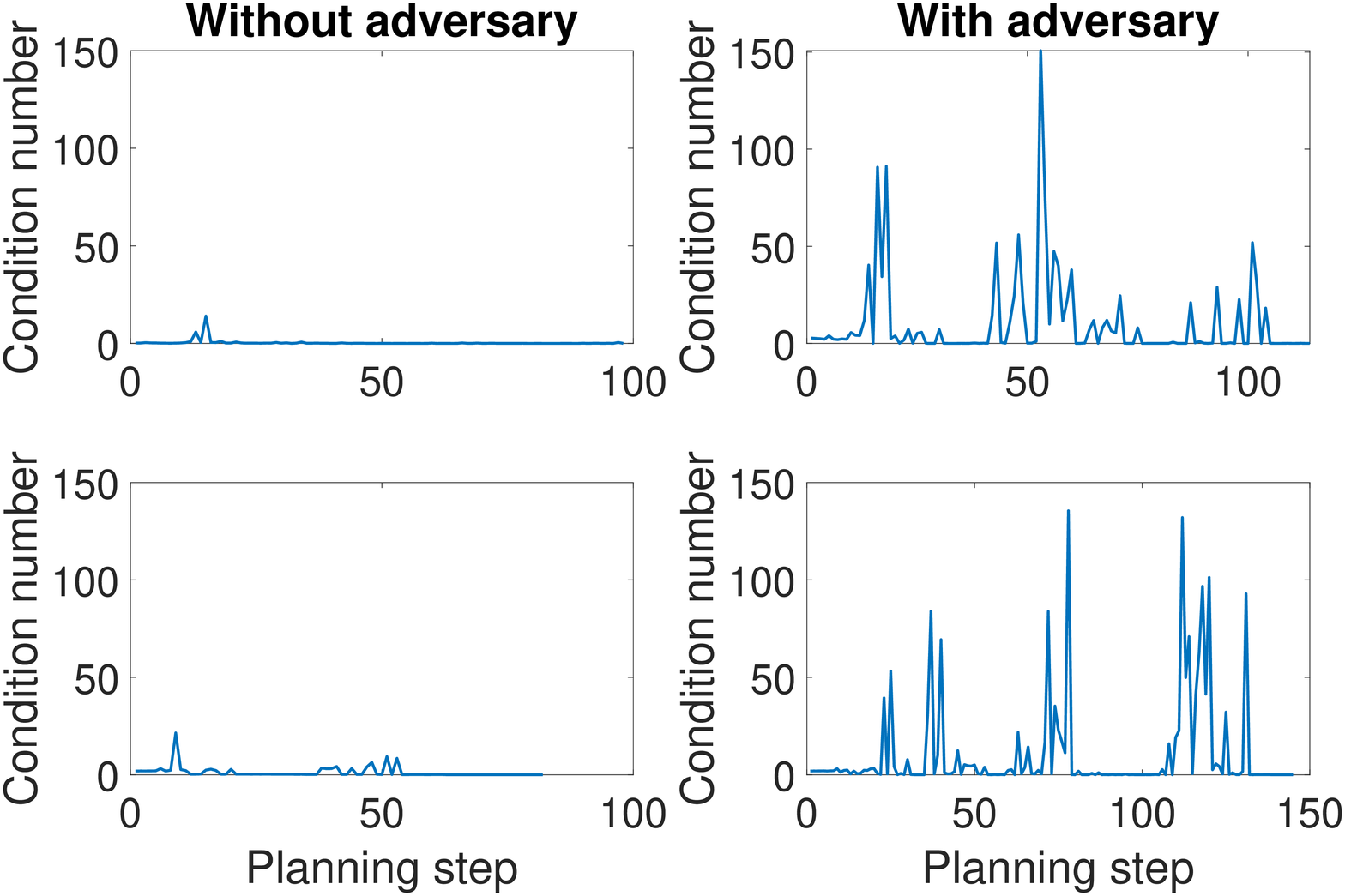}
    \caption{Loco planner: Scaled condition number of inverse Hessian - $\frac{\kappa}{100}$}
    \label{fig:cn}
  \end{subfigure}
  \quad
  \begin{subfigure}[t]{0.2\textwidth}
    \includegraphics[width=\textwidth]{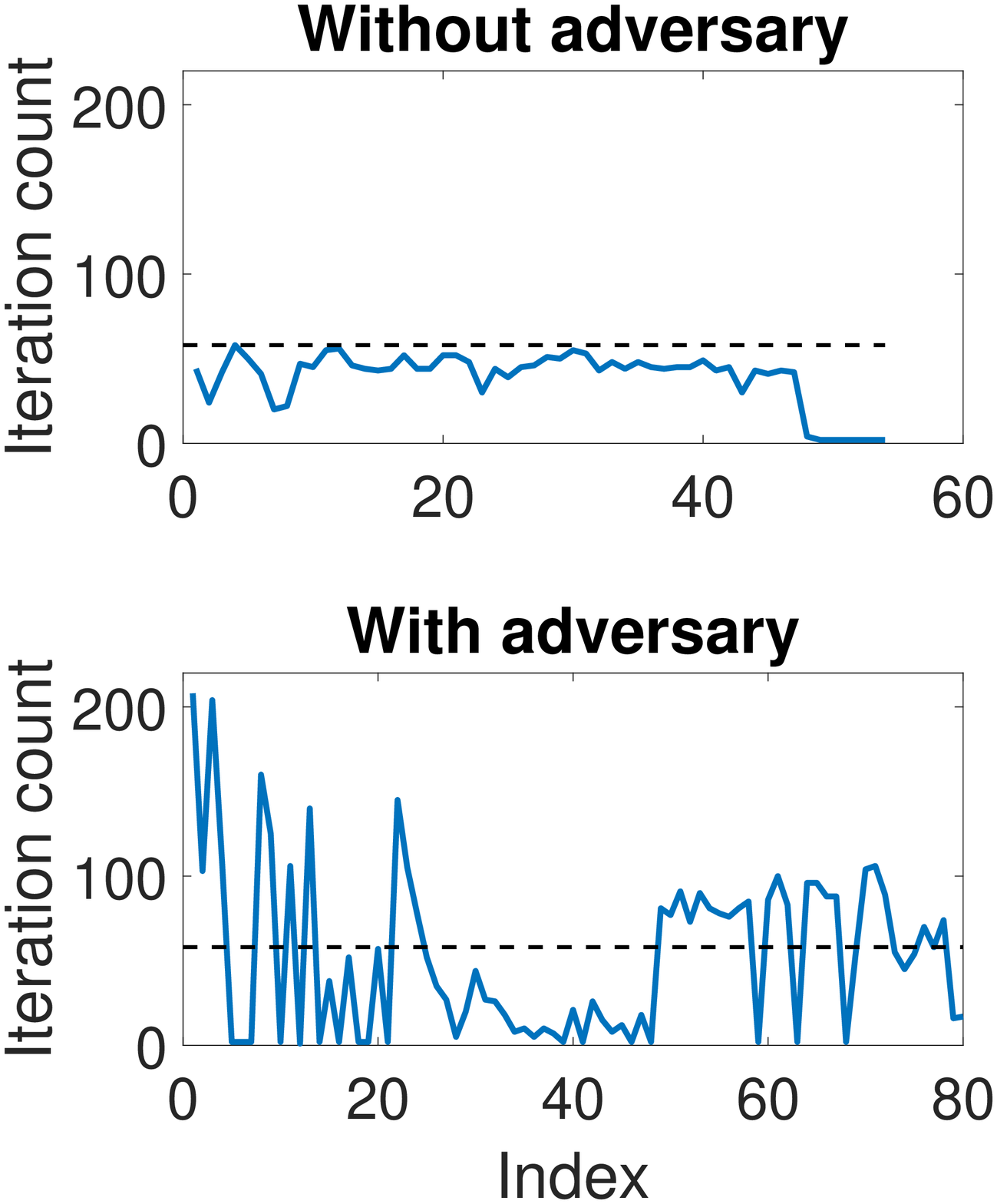}
    \caption{Loco: Optimizer iterations during a trial}
    \label{fig:loco_its}
  \end{subfigure}
  \quad
   \begin{subfigure}[t]{0.2\textwidth}
    \includegraphics[width=\textwidth]{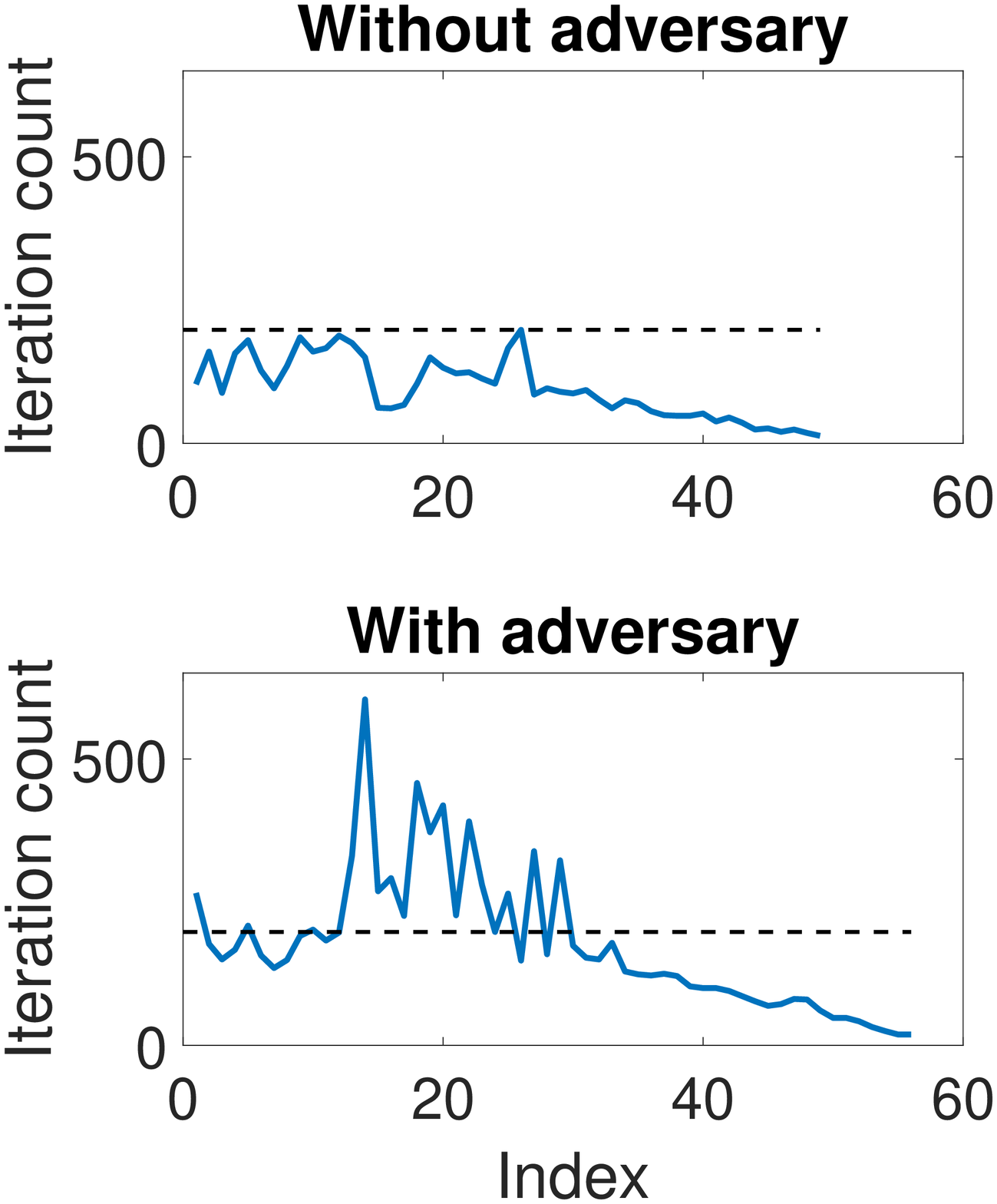}
    \caption{Fast: Optimizer iterations during a trial}
    \label{fig:fast_its}
  \end{subfigure}
  \caption{Effect of adversary on optimizers: Changes in condition number indicate a degradation of problem structure, and replanning requires higher number of iterations when adversary is present. Dashed line indicates maximum number of iterations seen with no adversary.}
  \label{fig:loco_cn}
\end{figure*}

First, we conduct adversarial attacks against both planners in the \emph{sparse} environment. A depiction of the effect of an adversarial attack can be seen in the comparison between \ref{fig:normalplan} and \ref{fig:advplan}. The replanned trajectories from start to goal in the absence of an adversary show a smooth profile but in the presence of an adversary, some of the candidate trajectories become unsafe, indicating failures within the planner.

We consistently observe a higher number of iterations required by the optimizer when the adversary is present compared to when it's not; for both the default and conservative configurations. Figures \ref{fig:fastsparse} and \ref{fig:locosparse} shows a comparison of the average number of iterations per trial; and also the maximum iteration count seen during a trial, averaged over 100 trials. Due to the large amount of open space, we find that collisions with the map to be rare.

\begin{table}
\footnotesize
\centering
\begin{adjustbox}{max width=0.5\textwidth}
\begin{tabular}{c|c|cc|ccc} 
\toprule
\multirow{2}{*}{Planner} & \multirow{2}{*}{Configuration} & \multicolumn{2}{c|}{No Adversary}     & \multicolumn{3}{c}{With Adversary}                                   \\ 
\cline{3-7}
                                     &                                & \multicolumn{1}{c|}{Success} & Coll-M & \multicolumn{1}{c|}{Success} & \multicolumn{1}{c|}{Coll-M} & Coll-A  \\ 
\hline
\multirow{2}{*}{Loco - BFGS}         & Default                        & 98\%                         & 2\%    & 10\%                         & 90\%                        & 0\%     \\ 
\cline{2-2}
                                     & Conservative                   & 99\%                         & 1\%    & 53\%                         & 45\%                        & 2\%     \\ 
\cline{1-2}
\multirow{2}{*}{Fast - LBFGS}        & Default                        & 84\%                         & 16\%   & 11\%                         & 86\%                        & 3\%     \\ 
\cline{2-2}
                                     & Conservative                   & 100\%                        & 0\%    & 52\%                         & 46\%                        & 2\%     \\ 
\cline{1-2}
\multirow{2}{*}{Fast - SLSQP}        & Default                        & 88\%                         & 12\%   & 16\%                         & 82\%                        & 2\%     \\ 
\cline{2-2}
                                     & Conservative                   & 100\%                        & 0\%    & 44\%                         & 48\%                        & 8\%     \\
\bottomrule
\end{tabular}
\end{adjustbox}
\caption{Effect of adversary on target planning outcome in the \emph{dense} environment, over 100 trials. Success: No collisions, Coll-M: Target collided with the map, Coll-A: Target collided with the adversary.}
\end{table}

\begin{figure}[!t]
        \centering
        \begin{subfigure}[b]{0.19\textwidth}
            \includegraphics[width=\textwidth]{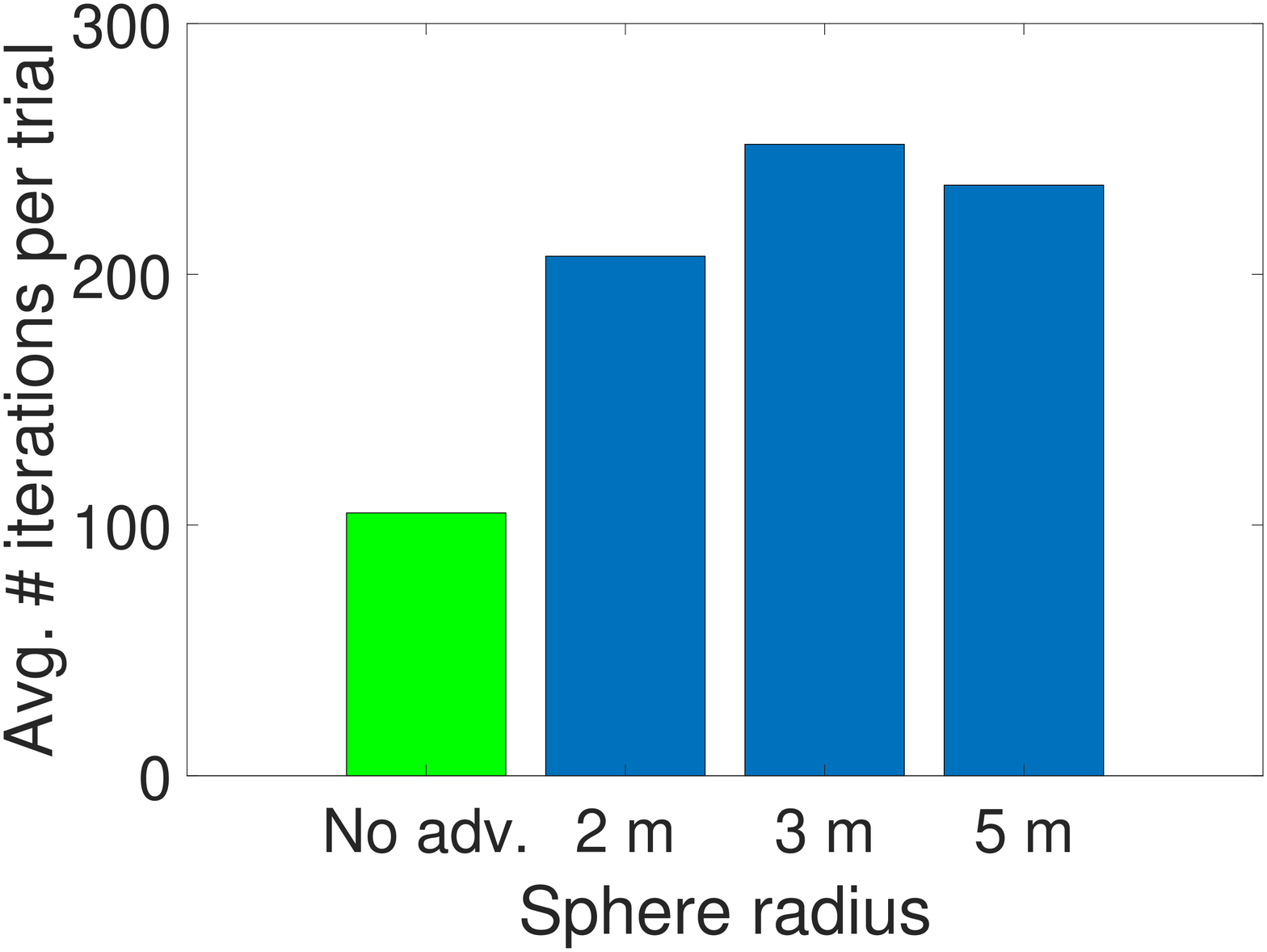}
            \caption{Fast - avg. iteration count}
            \label{fig:fast_sphere_avg}
        \end{subfigure}
        \begin{subfigure}[b]{0.19\textwidth} 
            \includegraphics[width=\textwidth]{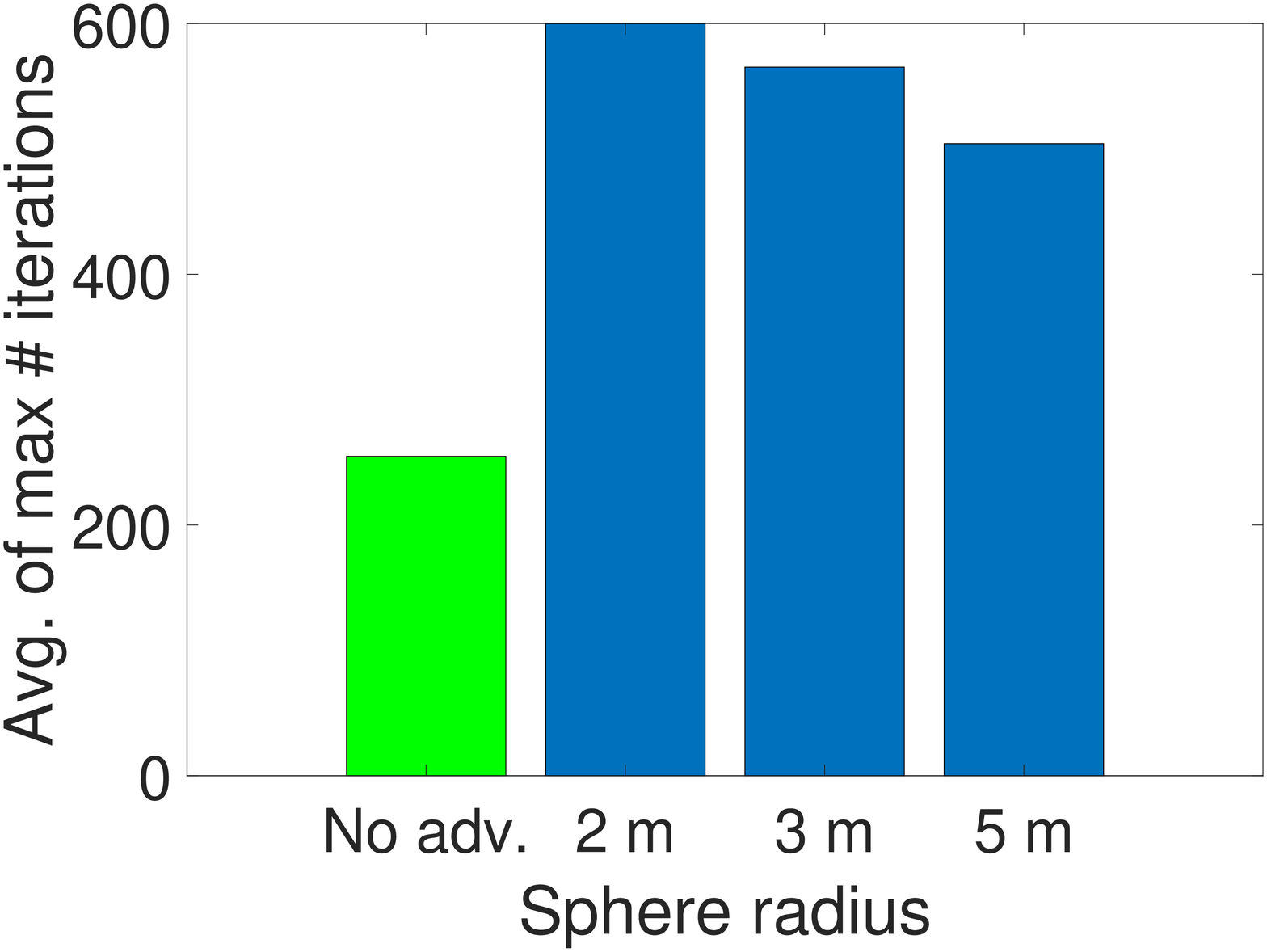}
            \caption{Fast - max. iterations}
            \label{fig:fast_sphere_max}
        \end{subfigure}
        \begin{subfigure}[t]{0.19\textwidth} 
            \includegraphics[width=\textwidth]{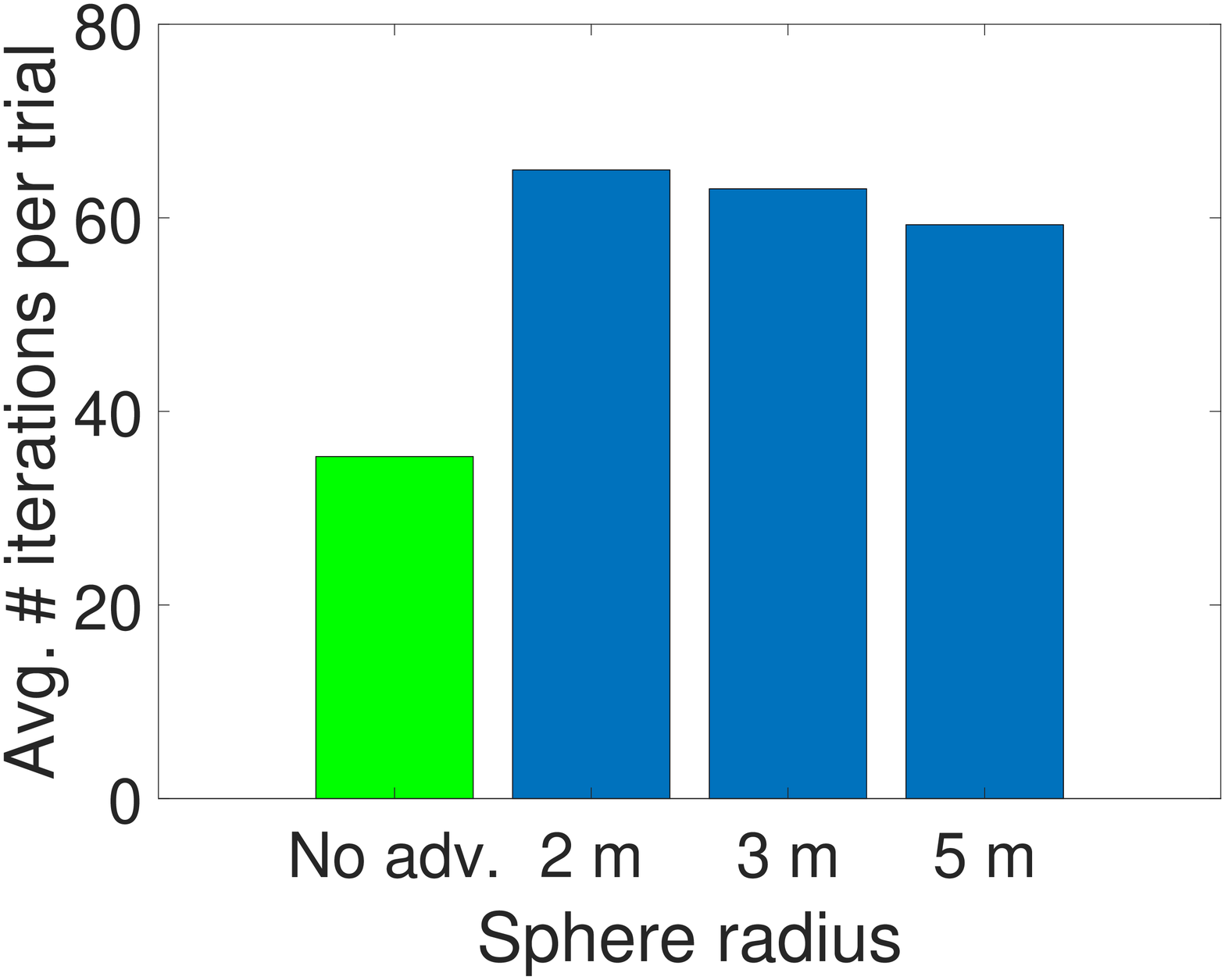}
            \caption{Loco - avg. iteration count}
            \label{fig:loco_sphere_avg}
        \end{subfigure}
        \begin{subfigure}[t]{0.19\textwidth}  
            \includegraphics[width=\textwidth]{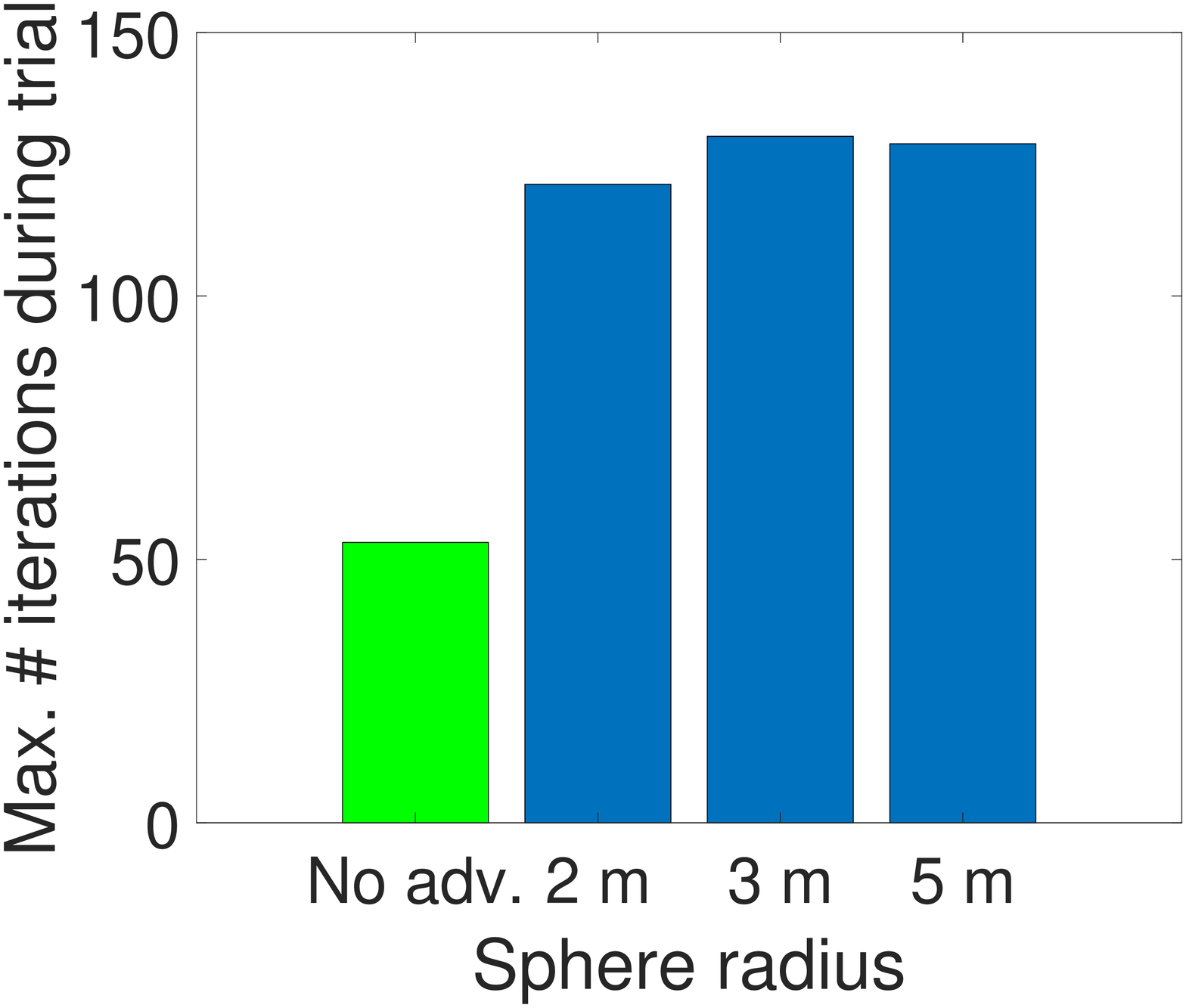}
            \caption{Loco - max. iterations}
            \label{fig:loco_sphere_max}
        \end{subfigure}
        \caption{Effect of obstacle proximity on the average iteration count of target optimizer, and the maximum number of iterations seen over all replanning steps in a trial: averaged over 10 random trials.} 
        \label{fig:sphere_prox}
    \end{figure}

We also evaluate the effect of the proximity of the adversary to the effort required by the optimizer. This is to demonstrate that the adversarial attack is not about simply being too close to the target, but whether the attack is able to maintain its effect (due to the weaknesses in the structure of the optimization problem) even at higher distances. This can be achieved by limiting search criterion by setting distance bounds either for the Bayesian optimization or the heuristic approximation. Figure \ref{fig:sphere_prox} shows that the effect on the target's optimizer stays consistent over different radii, thus demonstrating that the adversary can affect the target optimizer from a safe distance. 

Moving to the \emph{dense} environment, we notice both a high number of iterations when the adversarial attacks are carried out (Fig. \ref{fig:fastdense}, Fig. \ref{fig:locodense}), as well as failures to plan collision-free paths (Table 1). The \textit{default} setting was more failure-prone for all three planners with a 10-16\% success rate. The \textit{conservative} setting improved the general success rate, but with the adversary present, we observe a 40-50\% failure rate. The complex layout of the map coupled with the adversarial attack poses a hard problem structure, forcing the optimizers to fall into local minima and fail often.

We also compare these results with the behavior of the planner when the adversary robot is present, but moves in a random straight line path rather than occupying adversarial attack locations. Over 100 trials in the \emph{dense} map, this test shows the target reaching its goal 90\% of the time. Hence, the adversarial attack has a higher and more consistent effect on planner failure than a random moving obstacle. As another special case, we consider the scenario where the safety radius is ignored. This allows the adversary to collide with the robot directly. We simulate this scenario in the \emph{dense} map, and notice over 100 trials, the target collides with the adversary 77\% of the time, with the map 21\% of the time, and succeeds only 2\% of the time. Compared to table 1, this is a higher rate of failure, but we consider this an undesirable scenario as it results in the adversary getting destroyed in the process. 

For a better picture into the internal configuration of the planners, we log the maximum condition number observed during a planning trial in the Loco planner. Figure \ref{fig:cn} shows that during adversarial attacks, the condition number occasionally rises to very high values during replanning, indicating that the problem becomes ill-posed with a bad eigenstructure; compared to a cleaner profile when no adversary is present, indicating a stabler problem. A closer look into the profiles of iterations during a single trial are shown in Figs. \ref{fig:loco_its} and \ref{fig:fast_its}. In this sample trial, if we consider the maximum number of iterations seen without the adversary as a limit, Loco and Fast exceed this limit in 40\% and 30\% of the replanning steps respectively when the adversary is present. 

%% file: sections/conclusions.tex
\section{Conclusions}
\label{sec:conclusion}

In this paper, we present an approach to perform black-box adversarial attacks on optimization based collision-avoidance planners. We build this upon the idea that an adversary present in the same environment as a target robot can exploit the fact that its state forms part of the cost function the target is attempting to optimize. This allows the adversary to generate and occupy certain states that cause the optimization problem structure to degrade by affecting the condition number. We use Bayesian optimization over a surrogate Gaussian process prior to model behavior of a collision avoidance planner to result in these adversarial attacks, which we then approximate with a universal heuristic. We show that this heuristic is successful in affecting performance of multiple algorithms, either driving the internal optimizers to return invalid paths, or increasing the convergence time. The adversarial attack can thus be considered to be a special case of an obstacle's state in the problem structure of the planner, thus making it practically feasible. We plan to investigate the effect of these attacks on planners with hard constraints or certifiable safety as part of future efforts. 

We note here that the focus of the adversarial attacks is primarily to demonstrate the effect of dynamic adversarial configurations on iterative optimization. Through careful tuning and environment-specific modifications, a particular planner can be made to perform better against the dynamic obstacles, but usually at the expense of a larger number of iterations. We point out the effects of adversarial attacks not as weaknesses of these particular planning algorithms used, but of the iterative optimization scheme itself. These adversarial attacks could be mitigated by relying on non-iterative schemes for planning, such as through selection of motion/trajectory primitives or neural network models.

%% file: root.bbl
\begin{thebibliography}{10}
\providecommand{\url}[1]{#1}
\csname url@rmstyle\endcsname
\providecommand{\newblock}{\relax}
\providecommand{\bibinfo}[2]{#2}
\providecommand\BIBentrySTDinterwordspacing{\spaceskip=0pt\relax}
\providecommand\BIBentryALTinterwordstretchfactor{4}
\providecommand\BIBentryALTinterwordspacing{\spaceskip=\fontdimen2\font plus
\BIBentryALTinterwordstretchfactor\fontdimen3\font minus
  \fontdimen4\font\relax}
\providecommand\BIBforeignlanguage[2]{{%
\expandafter\ifx\csname l@#1\endcsname\relax
\typeout{** WARNING: IEEEtran.bst: No hyphenation pattern has been}%
\typeout{** loaded for the language `#1'. Using the pattern for}%
\typeout{** the default language instead.}%
\else
\language=\csname l@#1\endcsname
\fi
#2}}

\bibitem{ratliff2009chomp}
N.~Ratliff, M.~Zucker, J.~A. Bagnell, and S.~Srinivasa, ``Chomp: Gradient
  optimization techniques for efficient motion planning,'' in \emph{2009 IEEE
  International Conference on Robotics and Automation}.\hskip 1em plus 0.5em
  minus 0.4em\relax IEEE, 2009, pp. 489--494.

\bibitem{Oleynikova2016}
H.~{Oleynikova}, M.~{Burri}, Z.~{Taylor}, J.~{Nieto}, R.~{Siegwart}, and
  E.~{Galceran}, ``Continuous-time trajectory optimization for online uav
  replanning,'' in \emph{2016 IEEE/RSJ International Conference on Intelligent
  Robots and Systems (IROS)}, 2016, pp. 5332--5339.

\bibitem{zhou2019robust}
B.~Zhou, F.~Gao, L.~Wang, C.~Liu, and S.~Shen, ``Robust and efficient quadrotor
  trajectory generation for fast autonomous flight,'' \emph{IEEE Robotics and
  Automation Letters}, vol.~4, no.~4, pp. 3529--3536, 2019.

\bibitem{goodfellow2014explaining}
I.~J. Goodfellow, J.~Shlens, and C.~Szegedy, ``Explaining and harnessing
  adversarial examples,'' \emph{arXiv preprint arXiv:1412.6572}, 2014.

\bibitem{szegedy2013intriguing}
C.~Szegedy, W.~Zaremba, I.~Sutskever, J.~Bruna, D.~Erhan, I.~Goodfellow, and
  R.~Fergus, ``Intriguing properties of neural networks,'' \emph{arXiv preprint
  arXiv:1312.6199}, 2013.

\bibitem{brown2017adversarial}
T.~B. Brown, D.~Man{\'e}, A.~Roy, M.~Abadi, and J.~Gilmer, ``Adversarial
  patch,'' \emph{arXiv preprint arXiv:1712.09665}, 2017.

\bibitem{lee2019physical}
M.~Lee and Z.~Kolter, ``On physical adversarial patches for object detection,''
  \emph{arXiv preprint arXiv:1906.11897}, 2019.

\bibitem{eykholt2018robust}
K.~Eykholt, I.~Evtimov, E.~Fernandes, B.~Li, A.~Rahmati, C.~Xiao, A.~Prakash,
  T.~Kohno, and D.~Song, ``Robust physical-world attacks on deep learning
  visual classification,'' in \emph{Proceedings of the IEEE Conference on
  Computer Vision and Pattern Recognition}, 2018, pp. 1625--1634.

\bibitem{sharif2016accessorize}
M.~Sharif, S.~Bhagavatula, L.~Bauer, and M.~K. Reiter, ``Accessorize to a
  crime: Real and stealthy attacks on state-of-the-art face recognition,'' in
  \emph{Proceedings of the 2016 acm sigsac conference on computer and
  communications security}, 2016, pp. 1528--1540.

\bibitem{601330}
U.~{Lindqvist} and E.~{Jonsson}, ``How to systematically classify computer
  security intrusions,'' in \emph{Proceedings. 1997 IEEE Symposium on Security
  and Privacy (Cat. No.97CB36097)}, 1997, pp. 154--163.

\bibitem{elsayed2020ddosnet}
M.~S. Elsayed, N.-A. Le-Khac, S.~Dev, and A.~D. Jurcut, ``Ddosnet: A
  deep-learning model for detecting network attacks,'' in \emph{2020 IEEE 21st
  International Symposium on" A World of Wireless, Mobile and Multimedia
  Networks"(WoWMoM)}.\hskip 1em plus 0.5em minus 0.4em\relax IEEE, 2020, pp.
  391--396.

\bibitem{doshi2018machine}
R.~Doshi, N.~Apthorpe, and N.~Feamster, ``Machine learning ddos detection for
  consumer internet of things devices,'' in \emph{2018 IEEE Security and
  Privacy Workshops (SPW)}.\hskip 1em plus 0.5em minus 0.4em\relax IEEE, 2018,
  pp. 29--35.

\bibitem{petsios2017slowfuzz}
T.~Petsios, J.~Zhao, A.~D. Keromytis, and S.~Jana, ``Slowfuzz: Automated
  domain-independent detection of algorithmic complexity vulnerabilities,'' in
  \emph{Proceedings of the 2017 ACM SIGSAC Conference on Computer and
  Communications Security}, 2017, pp. 2155--2168.

\bibitem{luo2019multi}
W.~Luo and A.~Kapoor, ``Multi-robot collision avoidance under uncertainty with
  probabilistic safety barrier certificates,'' \emph{arXiv preprint
  arXiv:1912.09957}, 2019.

\bibitem{ames2016control}
A.~D. Ames, X.~Xu, J.~W. Grizzle, and P.~Tabuada, ``Control barrier function
  based quadratic programs for safety critical systems,'' \emph{IEEE
  Transactions on Automatic Control}, vol.~62, no.~8, pp. 3861--3876, 2016.

\bibitem{wang2017safety}
L.~Wang, A.~D. Ames, and M.~Egerstedt, ``Safety barrier certificates for
  collisions-free multirobot systems,'' \emph{IEEE Transactions on Robotics},
  vol.~33, no.~3, pp. 661--674, 2017.

\bibitem{herbert2017fastrack}
S.~L. Herbert, M.~Chen, S.~Han, S.~Bansal, J.~F. Fisac, and C.~J. Tomlin,
  ``Fastrack: A modular framework for fast and guaranteed safe motion
  planning,'' in \emph{2017 IEEE 56th Annual Conference on Decision and Control
  (CDC)}.\hskip 1em plus 0.5em minus 0.4em\relax IEEE, 2017, pp. 1517--1522.

\bibitem{bajcsy2019efficient}
A.~Bajcsy, S.~Bansal, E.~Bronstein, V.~Tolani, and C.~J. Tomlin, ``An efficient
  reachability-based framework for provably safe autonomous navigation in
  unknown environments,'' in \emph{2019 IEEE 58th Conference on Decision and
  Control (CDC)}.\hskip 1em plus 0.5em minus 0.4em\relax IEEE, 2019, pp.
  1758--1765.

\bibitem{sadigh2015safe}
D.~Sadigh and A.~Kapoor, ``Safe control under uncertainty,'' \emph{arXiv
  preprint arXiv:1510.07313}, 2015.

\bibitem{dey2016fast}
\BIBentryALTinterwordspacing
D.~Dey, D.~Sadigh, and A.~Kapoor, ``Fast safe mission plans for autonomous
  vehicles,'' in \emph{Robotics: Science and Systems (RSS) 2016 Workshop on
  Task and Motion Planning}, June 2016. [Online]. Available:
  \url{https://www.microsoft.com/en-us/research/publication/fast-safe-mission-plans-autonomous-vehicles/}
\BIBentrySTDinterwordspacing

\bibitem{ulusoy2014incremental}
A.~Ulusoy, T.~Wongpiromsarn, and C.~Belta, ``Incremental controller synthesis
  in probabilistic environments with temporal logic constraints,'' \emph{The
  International Journal of Robotics Research}, vol.~33, no.~8, pp. 1130--1144,
  2014.

\bibitem{axelrod2018provably}
B.~Axelrod, L.~P. Kaelbling, and T.~Lozano-P{\'e}rez, ``Provably safe robot
  navigation with obstacle uncertainty,'' \emph{The International Journal of
  Robotics Research}, vol.~37, no. 13-14, pp. 1760--1774, 2018.

\bibitem{majumdar2017funnel}
A.~Majumdar and R.~Tedrake, ``Funnel libraries for real-time robust feedback
  motion planning,'' \emph{The International Journal of Robotics Research},
  vol.~36, no.~8, pp. 947--982, 2017.

\bibitem{kalakrishnan2011stomp}
M.~Kalakrishnan, S.~Chitta, E.~Theodorou, P.~Pastor, and S.~Schaal, ``Stomp:
  Stochastic trajectory optimization for motion planning,'' in \emph{2011 IEEE
  international conference on robotics and automation}.\hskip 1em plus 0.5em
  minus 0.4em\relax IEEE, 2011, pp. 4569--4574.

\bibitem{oleynikova2018safe}
H.~Oleynikova, Z.~Taylor, R.~Siegwart, and J.~Nieto, ``Safe local exploration
  for replanning in cluttered unknown environments for microaerial vehicles,''
  \emph{IEEE Robotics and Automation Letters}, vol.~3, no.~3, pp. 1474--1481,
  2018.

\bibitem{tordesillas2020faster}
J.~Tordesillas, B.~T. Lopez, M.~Everett, and J.~P. How, ``Faster: Fast and safe
  trajectory planner for flights in unknown environments,'' \emph{arXiv
  preprint arXiv:2001.04420}, 2020.

\bibitem{richter2016polynomial}
C.~Richter, A.~Bry, and N.~Roy, ``Polynomial trajectory planning for aggressive
  quadrotor flight in dense indoor environments,'' in \emph{Robotics
  Research}.\hskip 1em plus 0.5em minus 0.4em\relax Springer, 2016, pp.
  649--666.

\bibitem{nocedal2006numerical}
J.~Nocedal and S.~Wright, \emph{Numerical optimization}.\hskip 1em plus 0.5em
  minus 0.4em\relax Springer Science \& Business Media, 2006.

\bibitem{bertsekas1997nonlinear}
D.~P. Bertsekas, ``Nonlinear programming,'' \emph{Journal of the Operational
  Research Society}, vol.~48, no.~3, pp. 334--334, 1997.

\bibitem{nesterov2018lectures}
Y.~Nesterov, \emph{Lectures on convex optimization}.\hskip 1em plus 0.5em minus
  0.4em\relax Springer, vol. 137.

\bibitem{bottou2018optimization}
L.~Bottou, F.~E. Curtis, and J.~Nocedal, ``Optimization methods for large-scale
  machine learning,'' \emph{Siam Review}, vol.~60, no.~2, pp. 223--311, 2018.

\bibitem{mockus1978application}
J.~Mockus, V.~Tiesis, and A.~Zilinskas, ``The application of bayesian methods
  for seeking the extremum,'' \emph{Towards global optimization}, vol.~2, no.
  117-129, p.~2, 1978.

\bibitem{rasmussen2003gaussian}
C.~E. Rasmussen, ``Gaussian processes in machine learning,'' in \emph{Summer
  School on Machine Learning}.\hskip 1em plus 0.5em minus 0.4em\relax Springer,
  2003, pp. 63--71.

\bibitem{oleynikova2016signed}
H.~Oleynikova, A.~Millane, Z.~Taylor, E.~Galceran, J.~Nieto, and R.~Siegwart,
  ``Signed distance fields: A natural representation for both mapping and
  planning,'' in \emph{RSS 2016 Workshop: Geometry and Beyond-Representations,
  Physics, and Scene Understanding for Robotics}.\hskip 1em plus 0.5em minus
  0.4em\relax University of Michigan, 2016.

\end{thebibliography}
